%% file: iclr2017_conference.tex
\documentclass{article} 
\usepackage{iclr2017_conference,times}
\usepackage{hyperref}
\usepackage{url}

\usepackage{times}
\usepackage{url}
\usepackage{latexsym}
\usepackage{graphicx}
\usepackage{amsmath}
\usepackage{amsfonts}
\usepackage{amssymb}
\usepackage{dsfont}
\usepackage[english]{babel}
\usepackage{color}
\usepackage{booktabs}
\usepackage{multicol}

\title{Machine Comprehension Using Match-LSTM and Answer Pointer}
\author{Shuohang Wang\\
		School of Information Systems\\
		Singapore Management University\\ 
		\texttt{shwang.2014@phdis.smu.edu.sg}
	\And 
	Jing Jiang\\
	School of Information Systems\\
	Singapore Management University\\
	\texttt{jingjiang@smu.edu.sg}}
\def\ignore#1{}

\begin{document}

\maketitle
\begin{abstract}
Machine comprehension of text is an important problem in natural language processing. 
A recently released dataset, the Stanford Question Answering Dataset (SQuAD), offers a large number of real questions and their answers created by humans through crowdsourcing.
SQuAD provides a challenging testbed for evaluating machine comprehension algorithms, partly because compared with previous datasets, in SQuAD the answers do not come from a small set of candidate answers and they have variable lengths.
We propose an end-to-end neural architecture for the task.
The architecture is based on match-LSTM, a model we proposed previously for textual entailment, and Pointer Net, a sequence-to-sequence model proposed by \citet{vinyals2015pointer:NIPS2015} to constrain the output tokens to be from the input sequences.
We propose two ways of using Pointer Net for our task. 
Our experiments show that both of our two models substantially outperform the best results obtained by \citet{rajpurkar2016squad} using logistic regression and manually crafted features.
\end{abstract}

\input{intro}
\input{model}
\input{exp}

\input{related}
\input{conclusion}
\input{ack}

\bibliography{iclr2017_conference}
\bibliographystyle{iclr2017_conference}

\input{appendix}
\end{document}

%% file: intro.tex
\section{Introduction}
\label{sec:intro}

Machine comprehension of text is one of the ultimate goals of natural language processing.
While the ability of a machine to understand text can be assessed in many different ways, in recent years, several benchmark datasets have been created to focus on answering questions as a way to evaluate machine comprehension~\citep{richardsonmctest:EMNLP2013, hermann2015teaching:nips2015, hill2015goldilocks:ICLR2016, weston2015towards:ICLR2016, rajpurkar2016squad}.
In this setup, typically the machine is first presented with a piece of text such as a news article or a story.
The machine is then expected to answer one or multiple questions related to the text.

In most of the benchmark datasets, a question can be treated as a multiple choice question, whose correct answer is to be chosen from a set of provided candidate answers~\citep{richardsonmctest:EMNLP2013, hill2015goldilocks:ICLR2016}.  
Presumably, questions with more given candidate answers are more challenging. 
The Stanford Question Answering Dataset (SQuAD) introduced recently by \citet{rajpurkar2016squad} contains such more challenging questions whose
correct answers can be any sequence of tokens from the given text.
Moreover, unlike some other datasets whose questions and answers were created automatically in Cloze style~\citep{hermann2015teaching:nips2015, hill2015goldilocks:ICLR2016}, the questions and answers in SQuAD were created by humans through crowdsourcing, which makes the dataset more realistic.
Given these advantages of the SQuAD dataset, in this paper, we focus on this new dataset to study machine comprehension of text.
A sample piece of text and three of its associated questions are shown in Table~\ref{tab:example}.

Traditional solutions to this kind of question answering tasks rely on NLP pipelines that involve multiple steps of linguistic analyses and feature engineering, including syntactic parsing, named entity recognition, question classification, semantic parsing, etc.
Recently, with the advances of applying neural network models in NLP, there has been much interest in building end-to-end neural architectures for various NLP tasks, including several pieces of work on machine comprehension~\citep{hermann2015teaching:nips2015,hill2015goldilocks:ICLR2016,yin2016attention:NAACLWS2016,kadlec2016text:ACL2016,cui2016consensus:arxiv}.
However, given the properties of previous machine comprehension datasets, existing end-to-end neural architectures for the task either rely on the candidate answers~\citep{hill2015goldilocks:ICLR2016,yin2016attention:NAACLWS2016} or assume that the answer is a single token~\citep{hermann2015teaching:nips2015,kadlec2016text:ACL2016,cui2016consensus:arxiv}, which make these methods unsuitable for the SQuAD dataset. 
In this paper, we propose a new end-to-end neural architecture to address the machine comprehension problem as defined in the SQuAD dataset.

Specifically, observing that in the SQuAD dataset many questions are paraphrases of sentences from the original text, we adopt a match-LSTM model that we developed earlier for textual entailment~\citep{wang2015learning:NAACL2016}.
We further adopt the Pointer Net (Ptr-Net) model developed by \citet{vinyals2015pointer:NIPS2015}, which enables the predictions of tokens from the input sequence only rather than from a larger fixed vocabulary and thus allows us to generate answers that consist of multiple tokens from the original text.
We propose two ways to apply the Ptr-Net model for our task: a sequence model and a boundary model. 
We also further extend the boundary model with a search mechanism.
Experiments on the SQuAD dataset show that our two models both outperform the best performance reported by \citet{rajpurkar2016squad}.
Moreover, using an ensemble of several of our models, we can achieve very competitive performance on SQuAD.

Our contributions can be summarized as follows: 
(1) We propose two new end-to-end neural network models for machine comprehension, which combine match-LSTM and Ptr-Net to handle the special properties of the SQuAD dataset.
(2) We have achieved the performance of an exact match score of 67.9\% and an F1 score of 77.0\% on the unseen test dataset, which is much better than the feature-engineered solution~\citep{rajpurkar2016squad}.
Our performance is also close to the state of the art on SQuAD, which is 71.6\% in terms of exact match and 80.4\% in terms of F1 from Salesforce Research.
(3) Our further analyses of the models reveal some useful insights for further improving the method. Beisdes, we also made
our code available online~\footnote{ \url{https://github.com/shuohangwang/SeqMatchSeq}}.


\begin{table}[]
\centering
\small
\label{tab:example}
\begin{tabular}{ll}
\toprule
\multicolumn{2}{p{13cm}}{In 1870, Tesla moved to Karlovac, \textbf{to attend school at the Higher Real Gymnasium}, where he was profoundly influenced by a math teacher \textbf{Martin Sekuli{\'c}}. The classes were held in \textbf{German}, as it was a school within the Austro-Hungarian Military Frontier. Tesla was able to perform integral calculus in his head, which prompted his teachers to believe that he was cheating. He finished a four-year term in three years, graduating in 1873.} \\ 
\midrule
1. In what language were the classes given?                                                                                                                                                                                                                                                 & German                                                                                                                                                                                                                                                                                  \\
2. Who was Tesla's main influence in Karlovac?                                                                                                                                                                                                                                              & Martin Sekuli{\'c}
\\
3. Why did Tesla go to Karlovac?                                                                                                                                                                                                                                                            & attend school at the Higher Real Gymnasium                                                                                                                                                                                                                                             
\\ \bottomrule
\end{tabular}
\normalsize
\caption{A paragraph from Wikipedia and three associated questions together with their answers, taken from the SQuAD dataset. The tokens in bold in the paragraph are our predicted answers while the texts next to the questions are the ground truth answers.}
\end{table}

%% file: model.tex
\section{Method}

In this section, we first briefly review match-LSTM and Pointer Net.
These two pieces of existing work lay the foundation of our method.
We then present our end-to-end neural architecture for machine comprehension.

\subsection{Match-LSTM}

In a recent work on learning natural language inference, we proposed a match-LSTM model for predicting textual entailment~\citep{wang2015learning:NAACL2016}.
In textual entailment, two sentences are given where one is a premise and the other is a hypothesis.
To predict whether the premise entails the hypothesis, the match-LSTM model goes through the tokens of the hypothesis sequentially.
At each position of the hypothesis, attention mechanism is used to obtain a weighted vector representation of the premise.
This weighted premise is then to be combined with a vector representation of the current token of the hypothesis and fed into an LSTM, which we call the match-LSTM. 
The match-LSTM essentially sequentially aggregates the matching of the attention-weighted premise to each token of the hypothesis and uses the aggregated matching result to make a final prediction.

\subsection{Pointer Net}

\citet{vinyals2015pointer:NIPS2015} proposed a Pointer Network (Ptr-Net) model to solve a special kind of problems where we want to generate an output sequence whose tokens must come from the input sequence.
Instead of picking an output token from a fixed vocabulary, Ptr-Net uses attention mechanism as a pointer to select a position from the input sequence as an output symbol.
The pointer mechanism has inspired some recent work on language processing~\citep{gu2016copynet, kadlec2016text:ACL2016}.
Here we adopt Ptr-Net in order to construct answers using tokens from the input text.

\begin{figure}[]
\centering
\includegraphics[width=5.5in]{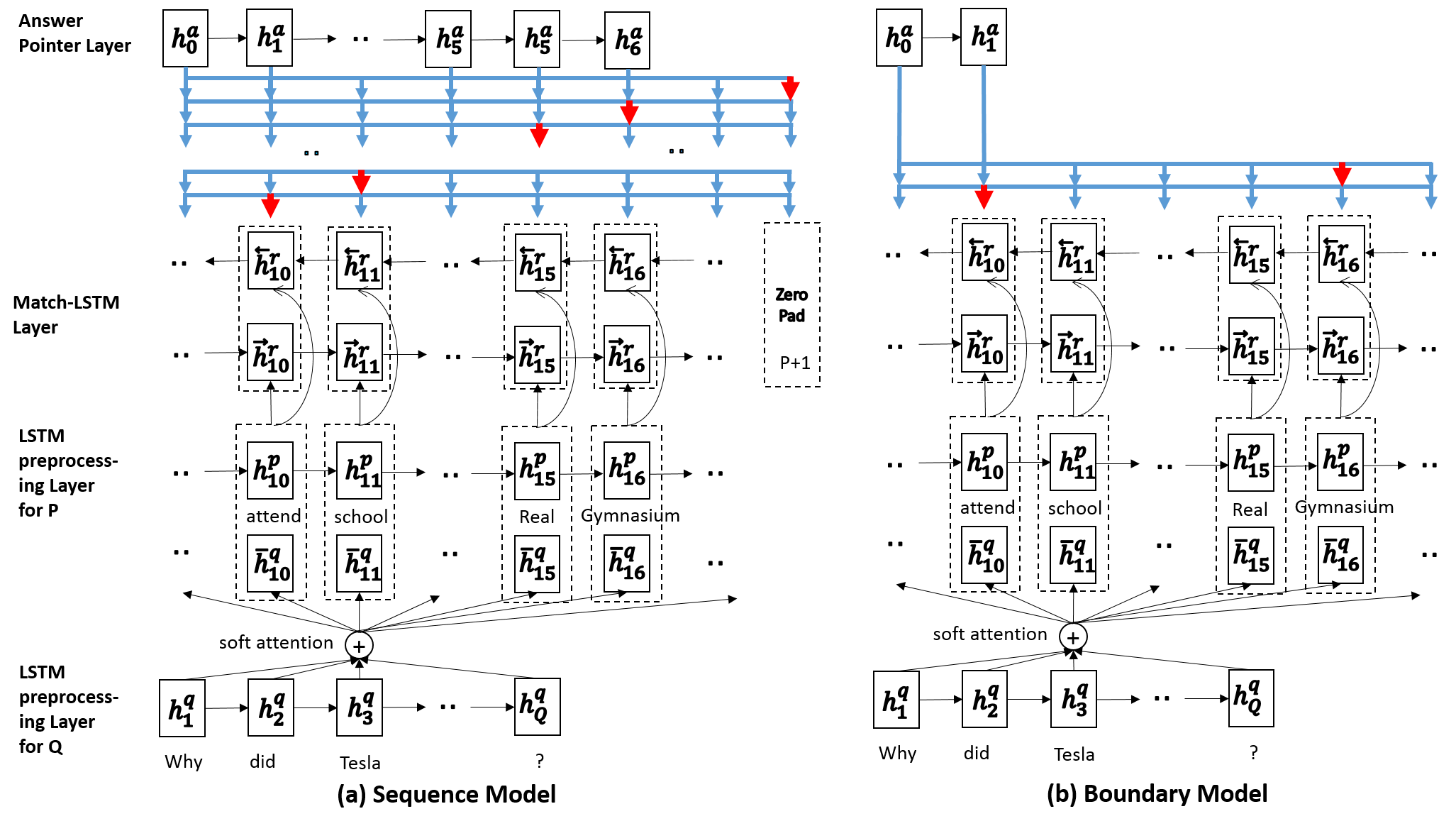}
\caption{An overview of our two models. 
Both models consist of an LSTM preprocessing layer, a match-LSTM layer and an Answer Pointer layer. For each match-LSTM in a particular direction, $\bar{h}^\text{q}_i$, which is defined as $\textbf{H}^\text{q} \alpha_i^{\intercal}$, is computed using the $\alpha$ in the corresponding direction, as described in either Eqn.~(\ref{eqn:alpha}) or Eqn.~(\ref{eqn:reverse_alpha}).}
\label{fig:model}
\end{figure}

\subsection{Our Method}

Formally, the problem we are trying to solve can be formulated as follows.
We are given a piece of text, which we refer to as a passage, and a question related to the passage.
The passage is represented by matrix $\mathbf{P} \in \mathbb{R}^{d \times P}$, where $P$ is the length (number of tokens) of the passage and $d$ is the dimensionality of word embeddings.
Similarly, the question is represented by matrix $\mathbf{Q} \in \mathbb{R}^{d \times Q}$ where $Q$ is the length of the question.
Our goal is to identify a subsequence from the passage as the answer to the question.

As pointed out earlier, since the output tokens are from the input, we would like to adopt the Pointer Net for this problem.
A straightforward way of applying Ptr-Net here is to treat an answer as a sequence of tokens from the input passage but ignore the fact that these tokens are consecutive in the original passage, because Ptr-Net does not make the consecutivity assumption.
Specifically, we represent the answer as a sequence of integers $\mathbf{a} = (a_1, a_2, \ldots)$, where each $a_i$ is an integer between 1 and $P$, indicating a certain position in the passage.

Alternatively, if we want to ensure consecutivity, that is, if we want to ensure that we indeed select a subsequence from the passage as an answer, we can use the Ptr-Net to predict only the start and the end of an answer.
In this case, the Ptr-Net only needs to select two tokens from the input passage, and all the tokens between these two tokens in the passage are treated as the answer.
Specifically, we can represent the answer to be predicted as two integers $\mathbf{a} = (a_\text{s}, a_\text{e})$, where $a_\text{s}$ an $a_\text{e}$ are integers between 1 and $P$.

We refer to the first setting above as a \emph{sequence} model and the second setting above as a \emph{boundary} model.
For either model, we assume that a set of training examples in the form of triplets $\{(\mathbf{P}_n, \mathbf{Q}_n, \mathbf{a}_n)\}_{n=1}^N$ are given.

\ignore{
Normally we could uniquely represent a subsequence by the positions of the start and the end of the subsequence in the original sequence.
In this paper, however, because we want to directly apply Pointer Net, we assume a relaxed setting in which we take tokens not necessarily consecutive in the given passage to form an answer.
We thus represent the answer as a sequence of integers $\mathbf{a} = (a_1, a_2, \ldots)$, where each $a_i$ is an integer between 1 and $P$, indicating a certain position in the passage.
We assume that a set of training examples in the form of triplets $\{(\mathbf{P}_n, \mathbf{Q}_n, \mathbf{a}_n)\}_{n=1}^N$ are given.
}

An overview of the two neural network models are shown in Figure~\ref{fig:model}.
Both models consist of three layers:
(1) An LSTM preprocessing layer that preprocesses the passage and the question using LSTMs.
(2) A match-LSTM layer that tries to match the passage against the question.
(3) An Answer Pointer (Ans-Ptr) layer that uses Ptr-Net to select a set of tokens from the passage as the answer.
The difference between the two models only lies in the third layer.

\noindent \textbf{LSTM Preprocessing Layer}

The purpose for the LSTM preprocessing layer is to incorporate contextual information into the representation of each token in the passage and the question.
We use a standard one-directional LSTM~\citep{hochreiter1997long} \footnote{As the output gates in the preprocessing layer affect the final performance little, we remove it in our experiments.} to process the passage and the question separately, as shown below:
\begin{equation}
\begin{matrix}
\mathbf{H}^{\text{p}} = \overrightarrow{\textit{LSTM}} (\mathbf{P}), & \mathbf{H}^{\text{q}} =\overrightarrow{\textit{LSTM}} (\mathbf{Q}).
\end{matrix}
\end{equation}
The resulting matrices $\mathbf{H}^{\text{p}} \in \mathbb{R}^{l \times P}$ and $\mathbf{H}^{\text{q}} \in \mathbb{R}^{l \times Q}$ are hidden representations of the passage and the question, where $l$ is the dimensionality of the hidden vectors.
In other words, the $i^{\text{th}}$ column vector $\mathbf{h}^{\text{p}}_i$ (or $\mathbf{h}^{\text{q}}_i$) in $\mathbf{H}^{\text{p}}$ (or $\mathbf{H}^{\text{q}}$) represents the $i^{\text{th}}$ token in the passage (or the question) together with some contextual information from the left.

\noindent \textbf{Match-LSTM Layer}

We apply the match-LSTM model~\citep{wang2015learning:NAACL2016} proposed for textual entailment to our machine comprehension problem by treating the question as a premise and the passage as a hypothesis.
The match-LSTM sequentially goes through the passage.
At position $i$ of the passage, it first uses the standard word-by-word attention mechanism to obtain 
attention weight vector $\overrightarrow{\alpha}_i \in \mathbb{R}^Q$ as follows:
\begin{eqnarray}
\nonumber
\overrightarrow{\mathbf{G}}_i & = & \text{tanh}(\mathbf{W}^{\text{q}} \mathbf{H}^{\text{q}} + (\mathbf{W}^{\text{p}} \mathbf{h}^{\text{p}}_i + \mathbf{W}^{\text{r}} \overrightarrow{\mathbf{h}}^{\text{r}}_{i-1} + \mathbf{b}^{\text{p}}) \otimes \mathbf{e}_Q ), \\
\label{eqn:alpha}
\overrightarrow{\alpha}_i & = & \text{softmax}(\mathbf{w}^\intercal \overrightarrow{\mathbf{G}}_i + b \otimes \mathbf{e}_Q),
\end{eqnarray}
where $\mathbf{W}^{\text{q}}, \mathbf{W}^{\text{p}}, \mathbf{W}^{\text{r}}\in \mathbb{R}^{l \times l}$, $\mathbf{b}^{\text{p}}, \mathbf{w} \in \mathbb{R}^l$ and $b \in \mathbb{R}$ are parameters to be learned, $\overrightarrow{\mathbf{h}}^{\text{r}}_{i-1} \in \mathbb{R}^l$ is the hidden vector of the one-directional match-LSTM (to be explained below) at position $i-1$, and the outer product $(\cdot \otimes \mathbf{e}_Q)$ produces a matrix or row vector by repeating the vector or scalar on the left for $Q$ times.

Essentially, the resulting attention weight $\overrightarrow{\alpha}_{i, j}$ above indicates the degree of matching between the $i^\text{th}$ token in the passage with the $j^\text{th}$ token in the question.
Next, we use the attention weight vector $\overrightarrow{\alpha}_i$ to obtain a weighted version of the question and combine it with the current token of the passage to form a vector $\overrightarrow{\mathbf{z}}_i$:
\ignore{
\begin{eqnarray}
\label{eqn:zh}
\overrightarrow{\mathbf{z}}_i  =  \begin{bmatrix}
\textbf{h}^\text{p}_i \\
\overrightarrow{\bar{\mathbf{h}}}^\text{q}_i
\end{bmatrix}, &
\overrightarrow{\bar{\mathbf{h}}}^\text{q}_i = \textbf{H}^\text{q} \overrightarrow{\alpha}_i^{\intercal}.
\end{eqnarray}
}

\begin{eqnarray}
\label{eqn:zh}
\overrightarrow{\mathbf{z}}_i  =  \begin{bmatrix}
\textbf{h}^\text{p}_i \\
\textbf{H}^\text{q} \overrightarrow{\alpha}_i^{\intercal}
\end{bmatrix}.
\end{eqnarray}

This vector $\overrightarrow{\mathbf{z}}_i$ is fed into a standard one-directional LSTM to form our so-called match-LSTM:
\begin{eqnarray}
\overrightarrow{\mathbf{h}}^\text{r}_i & = & \overrightarrow{\textit{LSTM}}(\overrightarrow{\mathbf{z}}_i, \overrightarrow{\mathbf{h}}^\text{r}_{i-1}),
\end{eqnarray}
where $\overrightarrow{\mathbf{h}}^\text{r}_i\in \mathbb{R}^l$.

We further build a similar match-LSTM in the reverse direction.
The purpose is to obtain a representation that encodes the contexts from both directions for each token in the passage.
To build this reverse match-LSTM, we first define
\begin{eqnarray}
\nonumber
\overleftarrow{\mathbf{G}}_i & = & \text{tanh}(\mathbf{W}^{\text{q}} \mathbf{H}^{\text{q}} + (\mathbf{W}^{\text{p}} \mathbf{h}^{\text{p}}_i + \mathbf{W}^{\text{r}} \overleftarrow{\mathbf{h}}^{\text{r}}_{i+1} + \mathbf{b}^{\text{p}}) \otimes \mathbf{e}_Q ), \\
\label{eqn:reverse_alpha}
\overleftarrow{\alpha}_i & = & \text{softmax}(\mathbf{w}^\intercal \overleftarrow{\mathbf{G}}_i + b \otimes \mathbf{e}_Q).
\end{eqnarray}
Note that the parameters here ($\mathbf{W}^{\text{q}}$, $\mathbf{W}^{\text{p}}$, $\mathbf{W}^{\text{r}}$, $\mathbf{b}^{\text{p}}$, $\mathbf{w}$ and $b$) are the same as used in Eqn.~(\ref{eqn:alpha}).
We then define $\overleftarrow{\mathbf{z}}_i$ in a similar way and finally define $\overleftarrow{\mathbf{h}}^\text{r}_i$ to be the hidden representation at position $i$ produced by the match-LSTM in the reverse direction.

Let $\overrightarrow{\mathbf{H}}^\text{r} \in \mathbb{R}^{l \times P}$ represent the hidden states $[\overrightarrow{\mathbf{h}}^\text{r}_{1}, \overrightarrow{\mathbf{h}}^\text{r}_{2}, \ldots, \overrightarrow{\mathbf{h}}^\text{r}_{P}]$ and $\overleftarrow{\mathbf{H}}^\text{r} \in \mathbb{R}^{l \times P}$ represent $[\overleftarrow{\mathbf{h}}^\text{r}_{1}, \overleftarrow{\mathbf{h}}^\text{r}_{2}, \ldots, \overleftarrow{\mathbf{h}}^\text{r}_{P}]$.
We define $\mathbf{H}^\text{r} \in \mathbb{R}^{2l \times P}$ as the concatenation of the two:
\begin{eqnarray}
\mathbf{H}^\text{r} & = & \begin{bmatrix}
\overrightarrow{\mathbf{H}}^\text{r} \\
\overleftarrow{\mathbf{H}}^\text{r}
\end{bmatrix}.
\end{eqnarray}

\noindent \textbf{Answer Pointer Layer}

The top layer, the Answer Pointer (Ans-Ptr) layer, is motivated by the Pointer Net introduced by \citet{vinyals2015pointer:NIPS2015}.
This layer uses the sequence $\mathbf{H}^\text{r}$ as input.
Recall that we have two different models: 
The \emph{sequence} model produces a sequence of answer tokens but these tokens may not be consecutive in the original passage.
The \emph{boundary} model produces only the start token and the end token of the answer, and then all the tokens between these two in the original passage are considered to be the answer.
We now explain the two models separately.

\noindent \textbf{The Sequence Model:} Recall that in the sequence model, the answer is represented by a sequence of integers $\mathbf{a} = (a_1, a_2, \ldots)$ indicating the positions of the selected tokens in the original passage.
The Ans-Ptr layer models the generation of these integers in a sequential manner.
Because the length of an answer is not fixed, in order to stop generating answer tokens at certain point, we allow each $a_k$ to take up an integer value between 1 and $P + 1$, where $P + 1$ is a special value indicating the end of the answer.
Once $a_k$ is set to be $P + 1$, the generation of the answer stops.

In order to generate the $k^\text{th}$ answer token indicated by $a_k$, first, the attention mechanism is used again to obtain an attention weight vector $\beta_k \in \mathbb{R}^{(P + 1)}$, where $\beta_{k, j}$ ($1 \leq j \leq P+1$) is the probability of selecting the $j^\text{th}$ token from the passage as the $k^\text{th}$ token in the answer, and $\beta_{k, (P + 1)}$ is the probability of stopping the answer generation at position $k$.
$\beta_k$ is modeled as follows:
\begin{eqnarray}
\mathbf{F}_k & = & \text{tanh}(\mathbf{V} \widetilde{\mathbf{H}}^\text{r} + (\mathbf{W}^\text{a} \mathbf{h}^\text{a}_{k-1} + \mathbf{b}^a )\otimes \mathbf{e}_{(P + 1)}), \\
\beta_k & = & \text{softmax}(\mathbf{v}^\intercal \mathbf{F}_k + c\otimes \mathbf{e}_{(P + 1)}),
\end{eqnarray}
where $\widetilde{\mathbf{H}}^\text{r} \in \mathbb{R}^{2l \times (P + 1)}$ is the concatenation of $\mathbf{H}^\text{r}$ with a zero vector, defined as $\widetilde{\mathbf{H}}^\text{r} = [\mathbf{H}^\text{r} ; \mathbf{0}]$, $\mathbf{V} \in \mathbb{R}^{l \times 2l}, \mathbf{W}^\text{a} \in \mathbb{R}^{l \times l}$, $\mathbf{b}^a, \mathbf{v} \in \mathbb{R}^l$ and $c \in \mathbb{R}$ are parameters to be learned, $(\cdot \otimes \mathbf{e}_{(P + 1)})$ follows the same definition as before, and $\mathbf{h}^\text{a}_{k-1} \in \mathbb{R}^l$ is the hidden vector at position $k - 1$ of an answer LSTM as defined below:
\begin{eqnarray}
\mathbf{h}^\text{a}_{k} = \overrightarrow{\textit{LSTM}}(\widetilde{\mathbf{H}}^\text{r} \mathbf{\beta}_k^{\intercal},\mathbf{h}^\text{a}_{k-1}).
\end{eqnarray}

We can then model the probability of generating the answer sequence as
\begin{eqnarray}
p(\mathbf{a} | \mathbf{H}^\text{r}) & = & \prod_{k} p(a_k | a_1, a_2, \ldots, a_{k-1}, \mathbf{H}^\text{r}),
\end{eqnarray}
and
\begin{eqnarray}
p(a_k = j | a_1, a_2, \ldots, a_{k-1}, \mathbf{H}^\text{r}) & = & \beta_{k, j}.
\end{eqnarray}

To train the model, we minimize the following loss function based on the training examples:
\begin{eqnarray}
-\sum_{n = 1}^N \log p(\mathbf{a}_n | \mathbf{P}_n, \mathbf{Q}_n).
\end{eqnarray}

\noindent \textbf{The Boundary Model:} The boundary model works in a way very similar to the sequence model above, except that instead of predicting a sequence of indices $a_1, a_2, \ldots$, we only need to predict two indices $a_\text{s}$ and $a_\text{e}$.
So the main difference from the sequence model above is that in the boundary model we do not need to add the zero padding to $\mathbf{H}^\text{r}$, and the probability of generating an answer is simply modeled as
\begin{eqnarray}
p(\mathbf{a} | \mathbf{H}^\text{r}) & = & p(a_\text{s} | \mathbf{H}^\text{r}) p(a_\text{e} | a_\text{s}, \mathbf{H}^\text{r}).
\end{eqnarray}

We further extend the boundary model by incorporating a search mechanism. 
Specifically, during prediction, we try to limit the length of the span and globally \textbf{search} the span with the highest probability computed by $p(\mathbf{a}_s)\times p(\mathbf{a}_e)$. 
Besides, as the boundary has a sequence of fixed number of values, \textbf{bi-directional Ans-Ptr} can be simply combined to fine-tune the correct span.

%% file: exp.tex
\section{Experiments}

In this section, we present our experiment results and perform some analyses to better understand how our models works.

\subsection{Data}

We use the Stanford Question Answering Dataset (SQuAD) v1.1 to conduct our experiments.
Passages in SQuAD come from 536 articles from Wikipedia covering a wide range of topics.
Each passage is a single paragraph from a Wikipedia article, and each passage has around 5 questions associated with it.
In total, there are 23,215 passages and 107,785 questions.
The data has been split into a training set (with 87,599 question-answer pairs), a development set (with 10,570 question-answer pairs) and a hidden test set.

\subsection{Experiment Settings}

We first tokenize all the passages, questions and answers.
The resulting vocabulary contains 117K unique words.
We use word embeddings from GloVe~\citep{pennington2014glove:emnlp2014} to initialize the model.
Words not found in GloVe are initialized as zero vectors.
The word embeddings are not updated during the training of the model.

The dimensionality $l$ of the hidden layers is set to be 150 or 300.
We use ADAMAX~\citep{kingma2014adam:iclr2015} with the coefficients $\beta_1=0.9$ and $\beta_2=0.999$ to optimize the model.
Each update is computed through a minibatch of 30 instances.
We do not use L2-regularization.

The performance is measured by two metrics: 
percentage of exact match with the ground truth answers, and word-level F1 score when comparing the tokens in the predicted answers with the tokens in the ground truth answers.
Note that in the development set and the test set each question has around three ground truth answers.
F1 scores with the best matching answers are used to compute the average F1 score.

\subsection{Results}

\begin{table}[]
\centering
	 \small
	
\begin{tabular}{lcccccc}
		\toprule
		\multicolumn{1}{c}{}      & \multicolumn{1}{c}{$l$} & \multicolumn{1}{c}{$|\theta|$} & \multicolumn{2}{c}{Exact Match}   & \multicolumn{2}{c}{F1}            \\ 
		&                       &                         & Dev             & Test            & Dev             & Test            \\ \midrule
		Random Guess              & -                     & 0                       & 1.1          & 1.3           & 4.1           & 4.3           \\
		Logistic Regression       & -                     & -                       & 40.0          & 40.4          & 51.0          & 51.0          \\
		DCR       & -                     & -                       &  62.5           & 62.5           & 71.2           & 71.0          \\ \midrule 
		Match-LSTM with Ans-Ptr (Sequence) & 150                   & 882K                    & 54.4 & - & 68.2 & - \\
		Match-LSTM with Ans-Ptr (Boundary) & 150                   & 882K                    & 61.1 & - & 71.2 & - \\
		Match-LSTM with Ans-Ptr (Boundary+Search) & 150                   & 882K                    & 63.0 & - & 72.7 & - \\
		Match-LSTM with Ans-Ptr (Boundary+Search) & 300                   & 3.2M                    & 63.1 & - & 72.7 & - \\
		Match-LSTM with Ans-Ptr (Boundary+Search+b) & 150                   & 1.1M                    & 63.4 & - & 73.0 & - \\
		Match-LSTM with Bi-Ans-Ptr (Boundary+Search+b) & 150                   & 1.4M                    & \textbf{64.1} & \textbf{64.7} & \textbf{73.9} & \textbf{73.7} \\ \midrule
		Match-LSTM with Ans-Ptr (Boundary+Search+en) & 150                   & 882K                    & \textbf{67.6} & \textbf{67.9} & \textbf{76.8} & \textbf{77.0} \\
		 \bottomrule
\end{tabular}
\normalsize
\caption{Experiment Results. Here ``Search'' refers to globally searching the spans with no more than 15 tokens, ``b'' refers to using bi-directional pre-processing LSTM, and ``en'' refers to ensemble method.}
\label{tab:results}
\end{table}

The results of our models as well as the results of the baselines given by \citet{rajpurkar2016squad} and \citet{Yu2015rank:arxiv} are shown in Table~\ref{tab:results}.
We can see that both of our two models have clearly outperformed the logistic regression model by \citet{rajpurkar2016squad}, which relies on carefully designed features.
Furthermore, our boundary model has outperformed the sequence model, achieving an exact match score of 61.1\% and an F1 score of 71.2\%. 
In particular, in terms of the exact match score, the boundary model has a clear advantage over the sequence model.
The improvement of our models over the logistic regression model shows that our end-to-end neural network models without much feature engineering are very effective on this task and this dataset.
Considering the effectiveness of boundary model, we further explore this model. 
Observing that most of the answers are the spans with relatively small sizes, we simply limit the largest predicted span to have no more than 15 tokens and conducted experiment with span searching
This resulted in 1.5\% improvement in F1 on the development data and that outperformed the DCR model~\citep{Yu2015rank:arxiv}, which also introduced some language features such as POS and NE into their model. 
Besides, we tried to increase the memory dimension $l$ in the model or add bi-directional pre-processing LSTM or add bi-directional Ans-Ptr. 
The improvement on the development data using the first two methods is quite small. 
While by adding Bi-Ans-Ptr with bi-directional pre-processing LSTM, we can get 1.2\% improvement in F1. 
Finally, we explore the ensemble method by simply computing the product of the boundary probabilities collected from 5 boundary models and then searching the most likely span with no more than 15 tokens.
This ensemble method achieved the best performance as shown in the table.

\subsection{Further Analyses}

\begin{figure}[]
	\centering
	\includegraphics[width=5.5in]{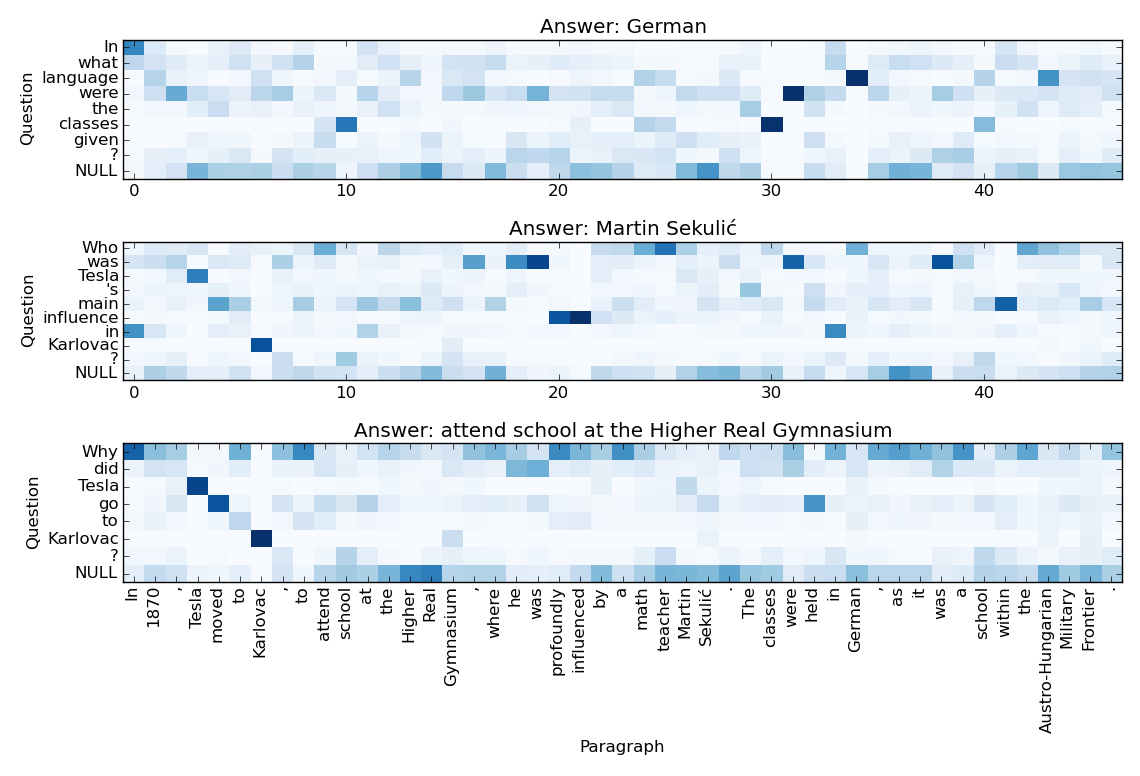}
	\caption{Visualization of the attention weights $\alpha$ for three questions associated with the same passage.  }
	\label{fig:alpha}
\end{figure}		

To better understand the strengths and weaknesses of our models, we perform some further analyses of the results below.

First, we suspect that longer answers are harder to predict.
To verify this hypothesis, we analysed the performance in terms of both exact match and F1 score with respect to the answer length on the development set.
For example, for questions whose answers contain more than 9 tokens, the F1 score of the boundary model drops to around 55\% and the exact match score drops to only around 30\%, compared to the F1 score and exact match score of close to 72\% and 67\%, respectively, for questions with single-token answers. And that supports our hypothesis.

Next, we analyze the performance of our models on different groups of questions.
We use a crude way to split the questions into different groups based on a set of question words we have defined, including ``what,'' ``how,'' ``who,'' ``when,''  ``which,'' ``where,'' and ``why.''
These different question words roughly refer to questions with different types of answers.
For example, ``when'' questions look for temporal expressions as answers, whereas ``where'' questions look for locations as answers.
According to the performance on the development data set, our models work the best for ``when'' questions.
This may be because in this dataset temporal expressions are relatively easier to recognize.
Other groups of questions whose answers are noun phrases, such as ``what'' questions, ``which'' questions and ``where'' questions, also get relatively better results.
On the other hand, ``why'' questions are the hardest to answer.
This is not surprising because the answers to ``why'' questions can be very diverse, and they are not restricted to any certain type of phrases.


Finally, we would like to check whether the attention mechanism used in the match-LSTM layer is effective in helping the model locate the answer.
We show the attention weights $\alpha$ in Figure~\ref{fig:alpha}.
In the figure the darker the color is the higher the weight is.
We can see that some words have been well aligned based on the attention weights.
For example, the word ``German'' in the passage is aligned well to the word ``language'' in the first question, and the model successfully predicts ``German'' as the answer to the question.
For the question word ``who'' in the second question, the word ``teacher'' actually receives relatively higher attention weight, and the model has predicted the phrase ``Martin Sekulic'' after that as the answer, which is correct.
For the last question that starts with ``why'', the attention weights are more evenly distributed and it is not clear which words have been aligned to ``why''.


%% file: related.tex
\section{Related Work}
Machine comprehension of text has gained much attention in recent years, and increasingly researchers are building data-drive, end-to-end neural network models for the task.
We will first review the recently released datasets and then some end-to-end models on this task.

\subsection{Datasets}
A number of datasets for studying machine comprehension were created in Cloze style by removing a single token from a sentence in the original corpus, and the task is to predict the missing word.
For example, \citet{hermann2015teaching:nips2015} created questions in Cloze style from CNN and Daily Mail highlights.
\citet{hill2015goldilocks:ICLR2016} created the Children's Book Test dataset, which is based on children's stories.
\citet{cui2016consensus:arxiv} released two similar datasets in Chinese, the People Daily dataset and the Children's Fairy Tale dataset.


Instead of creating questions in Cloze style, a number of other datasets rely on human annotators to create real questions.
\citet{richardsonmctest:EMNLP2013} created the well-known MCTest dataset
and \citet{MovieQA:cvpr2016} created the MovieQA dataset.
In these datasets, candidate answers are provided for each question.
Similar to these two datasets, the SQuAD dataset~\citep{rajpurkar2016squad} was also created by human annotators.
Different from the previous two, however, the SQuAD dataset does not provide candidate answers, and thus all possible subsequences from the given passage have to be considered as candidate answers.

Besides the datasets above, there are also a few other datasets created for machine comprehension, such as WikiReading dataset~\citep{hewlettwikireading:acl2016} and bAbI dataset~\citep{weston2015towards:ICLR2016}, but they are quite different from the datasets above in nature.

\subsection{End-to-end Neural Network Models for Machine Comprehension}

There have been a number of studies proposing end-to-end neural network models for machine comprehension.
A common approach is to use recurrent neural networks (RNNs) to process the given text and the question in order to predict or generate the answers~\citep{hermann2015teaching:nips2015}.
Attention mechanism is also widely used on top of RNNs in order to match the question with the given passage~\citep{hermann2015teaching:nips2015,chen2016thorough:ACL2016}.
Given that answers often come from the given passage, Pointer Network has been adopted in a few studies in order to copy tokens from the given passage as answers~\citep{kadlec2016text:ACL2016,trischler2016natural:emnlp2016}.
Compared with existing work, we use match-LSTM to match a question and a given passage, and we use Pointer Network in a different way such that we can generate answers that contain multiple tokens from the given passage.


Memory Networks~\citep{weston2014memory:ICLR2015} have also been applied to machine comprehension~\citep{sukhbaatar2015end:NIPS2015,kumar2015ask:ICML2016,hill2015goldilocks:ICLR2016},  but its scalability when applied to a large dataset is still an issue.
In this work, we did not consider memory networks for the SQuAD dataset.



%% file: conclusion.tex
\section{Conclusions}
In this paper,
We developed two models for the machine comprehension problem defined in the Stanford Question Answering (SQuAD) dataset, both making use of match-LSTM and Pointer Network.
Experiments on the SQuAD dataset showed that 
our second model, the boundary model, could achieve an exact match score of 67.6\% and an F1 score of 77\% on the test dataset, which is better than our sequence model and
\citet{rajpurkar2016squad}'s feature-engineered model.

In the future, we plan to look further into the different types of questions and focus on those questions which currently have low performance, such as the ``why' questions.
We also plan to test how our models could be applied to other machine comprehension datasets.


%% file: ack.tex
\section{Acknowledgments}
We thank Pranav Rajpurkar for testing our model on the hidden test dataset and Percy Liang for helping us with the Dockerfile for Codalab.

%% file: appendix.tex
\newpage
\appendix
\section{Appendix}
\begin{figure}[]
	\centering
	\includegraphics[width=5.5in]{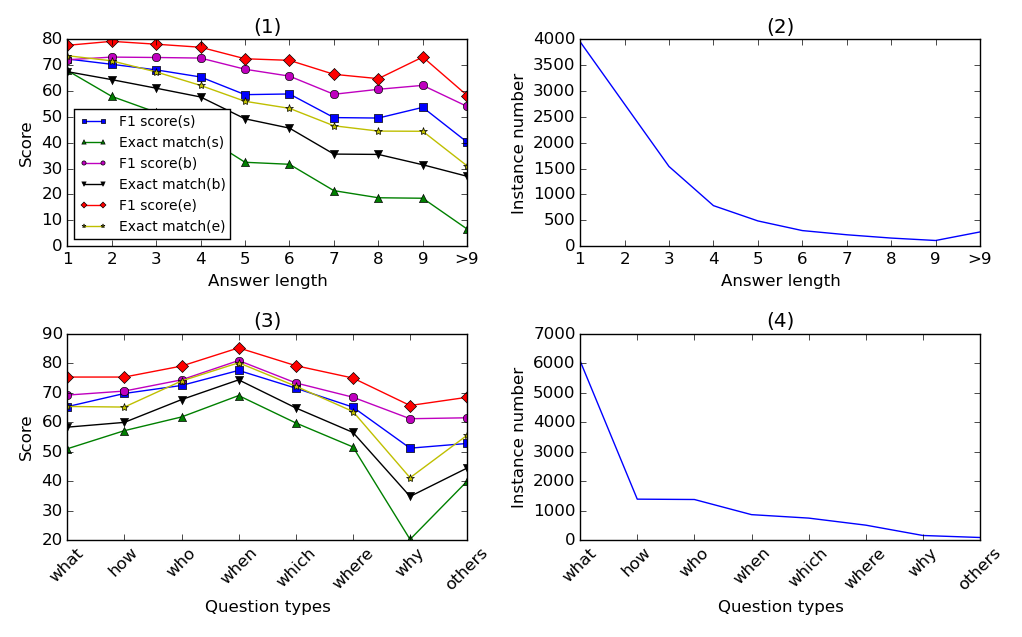}
	\caption{Performance breakdown by answer lengths and question types.  Top: Plot~(1) shows the performance of our two models (where \emph{s} refers to the sequence model , \emph{b} refers to the boundary model, and \emph{e} refers to the ensemble boundary model) over answers with different lengths. Plot~(2) shows the numbers of answers with different lengths. Bottom: Plot~(3) shows the performance our the two models on different types of questions. Plot~(4) shows the numbers of different types of questions.
	}
	\label{fig:ques}
\end{figure}

We show the performance breakdown by answer lengths and question types for our sequence model, boundary model and the ensemble model in Figure~\ref{fig:ques}.
